# Dense 3D Facial Reconstruction from a Single Depth Image in Unconstrained Environment

SHU ZHANG[1,2], HUI YU[2], TING WANG[3], JUNYU DONG[1,*], HONGHAI LIU[2]

[1]*Ocean University of China, Qingdao, China, 266100*

[2]*University of Portsmouth, Portsmouth, UK, PO1 2DJ*

[3]*Shandong University of Science & Technology, Qingdao, China, 266590*

*\*Corresponding author: dongjunyu@ouc.edu.cn*

**With the increasing demands of applications in virtual reality such as 3D films, virtual Human-Machine Interactions and virtual agents, the analysis of 3D human face analysis is considered to be more and more important as a fundamental step for those virtual reality tasks. Due to information provided by an additional dimension, 3D facial reconstruction enables aforementioned tasks to be achieved with higher accuracy than those based on 2D facial analysis. The denser the 3D facial model is, the more information it could provide. However, most existing dense 3D facial reconstruction methods require complicated processing and high system cost. To this end, this paper presents a novel method that simplifies the process of dense 3D facial reconstruction by employing only one frame of depth data obtained with an off-the-shelf RGB-D sensor. The experiments showed competitive results with real world data.**

## 1. INTRODUCTION

With the thriving developments of the applications in virtual reality such as 3D films, virtual Human-Machine Interaction (Essabbah et al. 2014; Hernoux and Christmann 2015) and virtual agents (Chen et al. 2016; Qu et al. 2013), 3D human face analysis is considered to be one of the most important researches (Lee et al. 2007) in this area. Compared to the analysis of 2D facial data, 3D facial analysis can achieve higher accuracy and efficiency than its equivalent 2D methods through the addition of another dimension. These advantages highly depend on the details of the 3D facial model. However, in practice, due to the limitations of the hardware and the structure of the scene such as depth shadowing, influence of the materials with reflection or refraction or infrared absorption in the scene, 3D information obtained from RGB-D data or many other reconstruction methods may often provide data that is insufficient for detailed facial analysis (Zhu et al. 2008). The resolutions of the 3D data are not good enough for detailed facial analysis. Moreover, high resolution 3D face reconstruction frequently requires particularly complicated preparation in the 3D acquisition process, or the usage of expensive equipment.

To tackle aforementioned problems, this paper presents a novel dense 3D facial reconstruction method that only utilizes one frame data from an off-the-shelf RGB-D sensor, even those device with the data in low resolutions. The reconstruction process is fully automatic without any manual intervention. The experiments reported in this paper demonstrate the competitive performance compared to other existing methods.

## 2. RELATED WORK

A number of 3D reconstruction methods exist in the literature and they can be classified into two groups: (a) the methods only utilizing RGB cameras, and (b) the methods based on dedicated depth sensors. Generally, the methods in the first group apply computer vision algorithms to multiple visual images for 3D recovery, for example, multi-view geometry, shape from shading. The most commonly used method with multi-view geometry is the stereo vision (Danescu et al. 2012). It utilizes two visual images, which are captured by two individual cameras placed with a fixed displacement, namely a baseline. A stereo rectification process is firstly conducted to reproject the image pair by two virtual cameras with the same focal length and no relative rotation to each other. Then the 3D structure can be achieved by analyzing the disparities from the matched 2D image features obtained with the epipolar constraints. With more than two images, the method called Structure from Motion (SfM) can recover the 3D structure of a scene as well as the poses of the cameras that capture these RGB images (Hartley and Zisserman 2003). Photometric Stereo (PMS) is another 3D reconstruction method based on shape from shading. By capturing a scene multiple times with different directional illuminations, PMS can recover a continuous depth map (Mecca et al. 2013). However, the camera should stay very still when capturing these images to make sure that the contents in these images are not moving.

The methods in the second group rely on dedicated hardware to recover 3D data. These hardware utilize techniques such as structure light (Nguyen et al. 2015) and time-of-flight (Chiabrando et al. 2009) to obtain 3D information of the scene. Normally, the structure light based methods are achieved by a structure light projector and a structure light receiver. A structured light projector projects a set of certain patterns onto the surfaces in the scene. The same pattern may change its appearance when projected onto the surfaces with different distances. By analyzing the pattern reflected from the scene, the 3D information of the scene can be calculated accordingly. For example, Microsoft's Kinect (Zhang 2012). The time-of-flight based methods scan the scene with multiple laser beams. By calculating the travel time of the laser beams, the 3D information of the scene can be obtained. For example, Niclass *et al.* (Niclass et al. 2014) introduced a depth sensor based on time-of-flight.

All these methods in the first group require capture of multiple images of the target either from different directions or using different illuminations. It is considered that these methods are neither able to generate 3D data with enough details for 3D facial analysis (e.g. stereo vision), nor be user friendly enough for human-robot interaction (e.g. photometric stereo). The performance of the methods in the second group are relatively more stable and more reliable. However, higher price is required if highly detailed 3D results are expected. With the studies of the advantages and disadvantages of the 3D reconstruction methods in both aforementioned groups, this paper presents a novel method suitable for detailed 3D facial recovery that can achieve a high data resolution while cost less for the equipment setup and preparation. The proposed method is achieved by the combination of the computer vision algorithms and the initial data from the off-the-shelf RGB-D sensor with a low price, for example, Microsoft's Kinect, Asus's Xtion Pro Live, Primesense's 3D Sensor and the Structure Sensor from Occipital, Inc., etc.

There is a range of methods that can reconstruct 3D human faces (Yu et al. 2012). However most of them either require complicated and special system setup, or come up with low accuracy results. For example, Bradley *et al.* (Bradley et al. 2010) presented a passive 3D facial capturing system, which was achieved by multi-view stereo. The results were obtained with high resolution. However, the data acquisition device is composed of multiple cameras and multiple illuminations from all directions. And the data were captured in a dark room to prevent the interference from the ambient light. Garcia *et al.* (Garcia et al. 2001) also proposed a low cost 3D face acquisition method. However, their method required to project a set of structure lights to form up a grid on the face directly, which was not user-friendly enough. The method presented by Hossain *et al.* (Hossain et al. 2007) utilized the stereo vision for 3D facial reconstruction. However, a man-made rectangular board is required behind the user's head, and four corners of this rectangle should be hand-coded with a red color before the process of stereo matching. Jo *et al.* (Jo et al. 2015) presented a 3D face reconstruction method. They combined the simplified 3DMM and the aforementioned SfM to achieve the 3D recovery. However, their method required manually annotated Facial Feature Points. Lee and Choi (Lee and Choi 2014) proposed a 3D face estimation method from a 2D frontal face image with an approach of the rank constraint relaxing. However, before their method is applied, the input image needs to be manually cropped according to the region of the face. Moreover, the resolution of the result was also fixed. Hwang *et al.* (Hwang et al. 2012) proposed a 3D face modelling method with continuous facial surface and texture. However, their method required two mirrors placed at both side of the face, and two chessboards were also required for calibrating the images from the mirrors, which made the system much complicated.

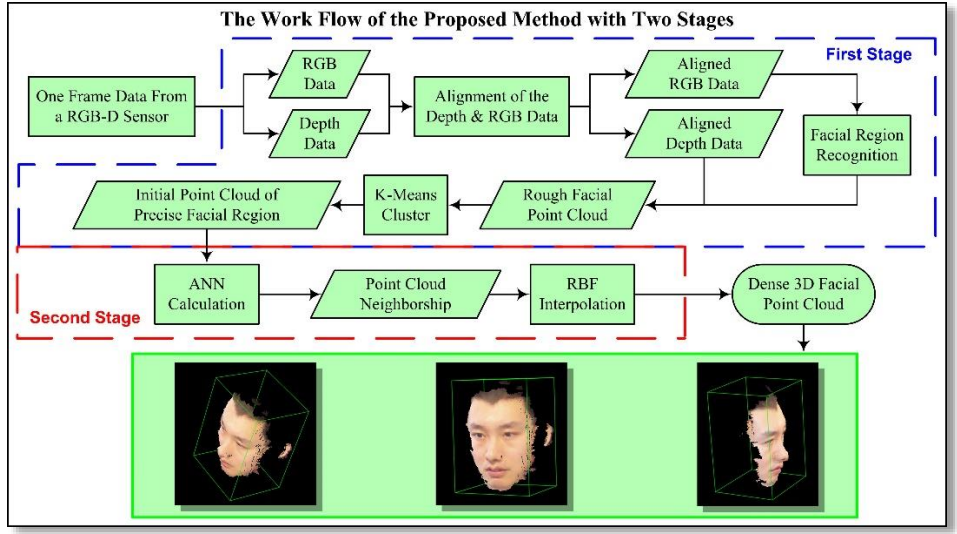

Fig. 1. Work flow of the proposed method.

Different from the aforementioned methods, the proposed 3D facial reconstruction method is fully automatic without any manual intervention. Moreover, a highly detailed 3D facial model can be obtained at a low system cost using our method, as shown in Figure 1. The process of the proposed method involves: (a) initial 3D point cloud acquisition; (b) precise face region extraction from the initial point cloud; (c) filling of the holes and blanks in the initial point cloud; (d) highly dense 3D facial point cloud generating using initial point cloud as seeds. Experiments of the proposed method illustrate encouraging results with real world data. The advantages of the proposed method in this paper are listed as follows.

1) Compared to most of the existing 3D facial reconstruction methods, the 3D recovery process of the proposed method is much simpler with minimal constraints. The only requirement for recovering the dense 3D facial point cloud using proposed method is one frame of the data from an off-the-shelf cheap RGB-D sensor without any additional setups.

2) The system cost of the proposed method is much lower than other existing 3D facial reconstruction methods. The only sensor needed in our method is a cheap RGB-D sensor with possibly low data resolution.

3) Compared to most of the existing methods, the whole 3D facial reconstruction process of the proposed method is fully automatic without any manual operation.

4) The output of the proposed method is a highly dense 3D facial point cloud with enriched facial details both in 3D structure and facial texture for further facial analysis.

The rest of the paper is organized as follow: the outline of the method is introduced in Section 3; In Section 4 and Section 5, two stages of the proposed method are described elaborately; the experiments with real world data are demonstrated and discussed in Section 6; the paper is concluded in Section 7.

## 3. OVERVIEW OF THE PROPOSED METHOD

This paper presents a novel method that can reconstruct a highly dense 3D human face. Furthermore, it only requires one frame data from an off-the-shelf RGB-D sensor, even those cheap sensors with low resolutions. The process is fully automatic without any manual intervention. The proposed method is divided into two main stages: (a) initial 3D facial point cloud acquisition and (b) dense facial point cloud propagation. The output of the first stage is a sparse 3D point cloud with precise facial region; the output of the final stage is a dense 3D facial point cloud with a smooth surface. Our method follows the outline below:

1) Extract one frame data from a RGB-D sensor, which contains an RGB image and a Depth image. With the parameters of the sensor,

registration between these two images is conducted to obtain 2D-3D correspondence between RGB image and Depth image.

2) Facial recognition is carried out on the RGB image. With the obtained facial region on RGB image and the 2D-3D correspondence between RGB and Depth data, the depth image is cropped to reserve the data that only belong to the facial area. To further refine the facial region on depth data, a k-means clustering algorithm is applied to remove the non-facial depth data. At the end of this step, an initial 3D point cloud can be obtained with a precise region of human face.

3) Approximate Nearest Neighbor (ANN) algorithm is applied to calculate the neighbor structure of the point cloud obtained in the previous step. With the neighborhood of the cloud, an interpolation based on Radial Basis Function (RBF) is employed to propagate a dense 3D facial point cloud from the initial point cloud. The detailed process is described in Section 4.

The work flow of the proposed method is shown in Figure 1, which describes the two main stages of the process. The first contribution of this paper is that the generation of dense 3D facial point cloud is fully automatic. No manual intervention is needed during the process. This is most suitable for the applications that require independent operations. The second contribution is that the highly dense facial point cloud can be obtained at a very low system cost. The only hardware needed in the proposed method is a cheap RGB-D sensor, even those with low resolutions. For example, the resolution of the depth data from the Kinect is only 640x480. To achieve full automation of 3D facial reconstruction, we innovatively introduce an automatic 3D facial region cropping scheme and an adoptive RBF based 3D interpolation process into the proposed method.

## 4. ACQUISITION OF INITIAL 3D FACIAL POINT CLOUD

The first stage of the proposed method is to acquire initial 3D point cloud for human face. In this paper, we utilize the Microsoft Kinect [7] as the input device to obtain the raw 3D data for demonstration. Kinect was originally designed for human machine interaction for game play. However, the features of the data captured by Kinect, especially 3D depth information, have also attracted the researchers in computer vision community. Similar with Xtion Pro Live, Primesense Sensor or the Structure Sensor, Kinect is also an RGB-D sensor that can simultaneously provide both RGB color and depth images. The RGB color image is captured by a built-in RGB digital camera inside the sensor. The depth data is achieved by an infrared laser projector and an infrared video camera mounted within the sensor.

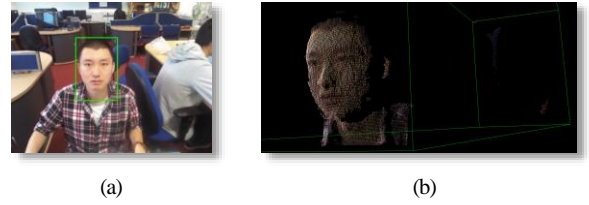

(a)          (b)

Fig. 2. 3D point cloud cropping by detected facial region. **(a)** The green rectangle is the facial region recognized in RGB image; **(b)** The 3D point cloud within the green frame is cropped to reserve the points that only belong to facial region detected in **(a)**.

Through the RGB image and depth image captured synchronically, it produces a spatial shift between these two images since the RGB camera and the infrared camera vary in spatial location (Han et al. 2013). Thus, we apply a registration process to the RGB image and depth image before we can utilize them. This alignment process can be achieved by taking into account of the constant distance between the RGB camera and the Infrared camera in the RGB-D device. With the knowledge of field of view (FOV) of the sensor, we can modify every pixel in the depth image accordingly to make them align with the pixel in RGB image. There is a built-in function in the Kinect SDK for the aforementioned registration process. However, we try to implement this process by our own to enhance the generalization and flexibility of the proposed method, for example, to enable the proposed method fully functional with the utilization of other RGB-D sensors rather than the Kinect.

With alignment, for every coordinate of 2D points in RGB image, we retrieve the corresponding 2D coordinate in depth image. Then coordinates of 3D points in the space can be obtained using (1).

$$\frac{x_p}{u-u_0} = \frac{y_p}{v-v_0} = \frac{z_p}{f} \qquad (1)$$

In (1), $(x_p, y_p, z_p)$ is the coordinate of the 3D point in the space, $(u, v)$ is the 2D coordinate of the point in the depth image, $(u_0, v_0)$ is the principle point of the infrared camera, and $f$ is the focal length of the infrared camera. With known value of $u$, $v$, $u_0$, $v_0$, $f$ and $z_p$, the 3D coordinate of $(x_p, y_p, z_p)$ can be retrieved.

After calculating the 3D points, the face detection is carried out on the aligned RGB image in order to filter out the 3D points to only reserve the points within the facial region, as shown in Figure 2. We implement the facial detection process using Haar Cascade Classifier (Wilson and Fernandez 2006). It trains the Haar feature classifiers for facial features such as eyes, mouth, and nose with two sets of images at first. One set contains pure human face without other objects while the other set contains one or more instance of objects. Then the facial region can be recognized in an RGB image with the trained classifiers. According to the registration between RGB image and 3D points, we only keep the 3D points inside the facial region, as shown in Figure 2 (b).

Though facial detection can remove most of the 3D points that do not belong to facial region, it hardly provides a precise region of a human face. Only a general area of the face is obtained. However, as it can be observed in Figure 2 (b), the 3D point cloud consists of several partitions, one of which is the real facial region. Therefore, to achieve a more accurate facial area in 3D point cloud, we apply a K-means clustering algorithm (Rao and Rao 2014) on the point cloud in Figure 2 (b) to divide it into several clusters, and only reserve the one that contains the center point of the facial region.

$$\begin{cases} \arg\min_{\mathbf{S}} \sum_{n=1}^{k} \sum_{x \in S_n} \|x - \mu_n\|^2 \\ \mathbf{S} = \{S_1, S_2, S_3, ..., S_n, ..., S_k\} \end{cases} \quad (2)$$

K-means algorithm achieves the clustering process by calculating the distance between points and centroids of the point groups, as shown in (2), where $\mu_n$ is the centroid coordinate of the cluster $S_n$. The initial centroids are selected randomly. After multiple iterations, a clustering solution $\mathbf{S}$ can be found when the similarity within a cluster is maximized and similarity inter-cluster is minimized. Since the facial detection has already provided a rough but tight facial region, we implement K-Means algorithm with 3 initial clusters, and reserve only the one containing the center point of the detected facial region. The result is shown in Figure 3. The facial region is much more precise than the one in Figure 2 (b), in which the 3D point cloud includes not only the facial points but also the points that belong to the background.

## 5. GENERATION OF DENSE FACIAL POINT CLOUD

As shown in Figure 3, the initial 3D facial point cloud is achieved in low quality with missing data and holes on the surface. Moreover, the point cloud is not dense enough to provide the details of the human face, which is essential for 3D facial analysis in some computer vision tasks. Therefore, the second stage of the proposed method is to apply an interpolation process to fill up the depth blank. A denser point cloud is propagated from the initial 3D facial points, which results in a smoother surface.

There is a range of interpolation algorithms aviable. Amidror Isaac (Amidror 2002) made a literature review for scattered data interpolation methods. He classified the interpolation methods into following groups: (a) the triangulation based methods, (b) the inverse distance weighted methods, (c) the radial basis fuction based methods, and (d) the natural neighbor interpolation methods. The interpolation process based on the triangulation is achieved by finding the central point of each trangular represented by the original data (Nielson et al. 1997). The calcualtion is local. However, it still requires to trangulate the given scattered point set as a preprocessing step. This process is sometimes computationally complex, especially for the dense interpolation which inserts more than one point within the origianl triangular. The inverse distance weighted methods, also known as Shepard methods, calculate the interpolated point values according to all the points in the origianl point cloud with the inverse distance of the points as the weights. However, since the Shepard methods are only sensitive to the distance, the accuracy of the results are easily affected by the overweight of the data clusters. The naturall neighbor interpolation methods are based on the Voronoi tesselation of the origianl point cloud (Sibson 1981).

In this paper, we choose the interpolation scheme based on the Radial Basis Function (RBF) (Fasshauer and McCourt 2012). The RBF based methods are considered to be some of the most elegant schemes, and often work very well (Franke and Nielson 1991). We treat the 3D point cloud as a set of spatial variables (X, Y, Z), where (X, Y) follows 2D coordinate, and (Z) is composed of a set of scalar variables, which are seated on the grid formed up by the aforementioned 2D coordinate. Accordingly, the interpolation process is to find the interpolated (Z) value in (X, Y) coordinate system. We apply RBF framework to perform interpolation with Gaussian

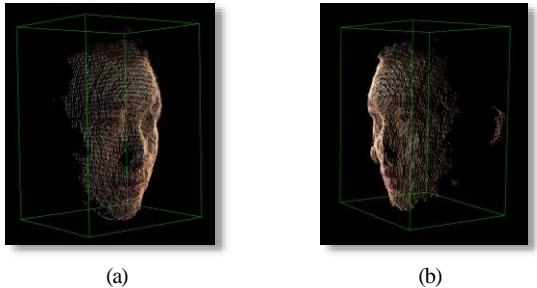

(a)　　　　　　　　　(b)

Fig. 3. The initial 3D facial point cloud obtained by the first stage of the proposed method. (a) and (b) are viewed from different angles. The facial region is achieved with more accuracy after applying k-means algorithm comparing to Figure 2 (b).

Kernel as the Basis Function. RBF model is built up by a set of Basis Functions with radial symmetry. It follows the form as in (3).

$$f(x) = \sum_{i=1}^{M}(w_i \cdot \varphi(\|x - c_i\|)) \quad (3)$$

$$F = \Phi W \quad (4)$$

where: $\varphi$ is a basis function, $w_i$ is the weight for each basis function, and $c_i$ is the interpolation center, which coincide with the original 3D point in the initial 3D facial point cloud obtained in last Section. The $c_i$ can be referred as the seed. Basis functions can be achieved with several forms, such as Gaussian basis function, Multi-Quadric Basis Function, Thin Plate Spline Basis Function, Cubic Basis Function and poly-harmonic Basis Functions. A value for an interpolated point in the coordinate system can be provided by the basis function according to the values of its neighboring seed points, and the distances between this interpolated point and its neighboring seed points. Therefore, as shown in (4), a linear system for all seed points $c_i$ can be found, where $F$ is a vector of the interpolated values, $W$ is a vector of the weights, and $\Phi$ is the interpolation matrix. The expansion of (4) is shown in (5). Obviously, with the knowledge of $f(c_i)$ for i=1,…,M, which is the value of each seed point, the weights $W$ can be derived from the system by (6).

$$\begin{bmatrix} f(x_1) \\ f(x_2) \\ \mathrm{M} \\ f(x_N) \end{bmatrix} = \begin{bmatrix} \varphi(\|x_1 - c_1\|) & \varphi(\|x_1 - c_2\|) & \mathrm{L} & \varphi(\|x_1 - c_M\|) \\ \varphi(\|x_2 - c_1\|) & \varphi(\|x_2 - c_2\|) & \mathrm{L} & \varphi(\|x_2 - c_M\|) \\ \mathrm{M} & \mathrm{M} & \mathrm{O} & \mathrm{M} \\ \varphi(\|x_N - c_1\|) & \varphi(\|x_N - c_2\|) & \mathrm{L} & \varphi(\|x_N - c_M\|) \end{bmatrix} \begin{bmatrix} w_1 \\ w_2 \\ \mathrm{M} \\ w_M \end{bmatrix} \quad (5)$$

$$W = \Phi^{-1} F \quad (6)$$

The Gaussian Basis Function is applied in our method as $\varphi$ expressed in (7) since Gaussian Basis Function is localized and compact, which means the value of the interpolated point is only sensitive to the values of its neighboring seed points. A distance of about $6 \cdot R_0$ will make the value almost equal zero with the Gaussian Basis Function, as shown in (7). This prevents interpolation from over-smooth which leads to losses of details on the 3D facial surface.

$$\varphi(r) = e^{\frac{-r^2}{2 \cdot \sigma^2}} = e^{\frac{-r^2}{R_0^2}} \quad (7)$$

As a result, the choice of $R_0$ is crucial to our method. If $R_0$ is too small, the interpolated points will prone to ground zero since there are not enough seed points within valid influential distance. The value will degrade rapidly when apart from the position of the seed point. If $R_0$ is too large, the calculation process becomes more time-consuming while interpolation performance doesn't improve too much. Our studies show that a better result emerges when $R_0$ is slightly larger than the average distance between original seed points in the coordinate system. Kd-Tree is commonly used to find a neighborhood of discrete points (Brown 2015). However, with the increment of the dimensionality and data complexity, the effectiveness of the classic Kd-Tree algorithm drops quickly. Therefore, we use Approximate Nearest Neighbors (ANN) algorithm (Har-Peled et al. 2012) instead of classic Kd-Tree (Exact Nearest Neighbors) algorithm to calculate the average distance of the points in initial 3D facial point cloud, and then form up the neighborhood of the points in point cloud for RBF based interpolation. Several ANN algorithm implementations exist, such as Best Bin First (Beis and Lowe 1997), Randomized Kd-Trees (Silpa-Anan and Hartley 2008), hierarchical k-Means Tree (Nister and Stewenius 2006), etc. In this paper, Randomized Kd-Trees (RKT) is employed for nearest neighbor search to construct the neighborhood of the initial 3D facial point cloud.

RKT is an extension of classic Kd-Tree algorithm. The classic Kd-Tree is a binary tree with k dimensional data. Similar to Binary Search Tree, A classic Kd-Tree is built by dividing the points in the space into several partitions according to multiple dimensions once at a time. For instance, in Figure 4, the blue vertical line on the left and blue circle in the first layer of the Kd-Tree on the right imply the division of the points along the first dimension. The green horizontal lines on the left and green circle in the second layer of the Kd-Tree on the right imply the division of the points divided by previous division along the second dimension. In each layer of the Kd-Tree, it chooses a dimension with max variance of the points, and divides the points along this dimension. Any point in the space can be located in one of the leaf nodes.

RKT is a forest with multiple Kd-trees initialized by different parameters, e.g., with different rotations on the data. The searches on these trees with different structures can be independent. Different from the classic Kd-Tree's strategy of choosing the dimension with max variance of the points to subdivide the data, RKT randomly chooses from a few of dimensions that have high data variance. This strategy enables the RKT with higher efficiency both in querying and backtracking.

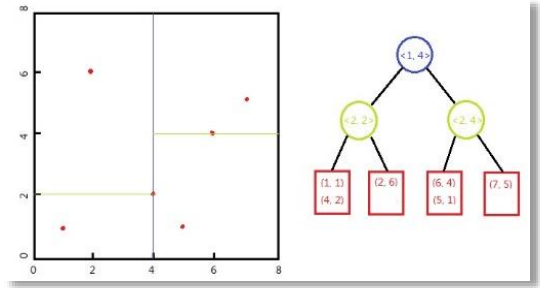

Fig. 4. Illustration of the Kd-Tree structure. The left part is the distribution of the points in 2D; the right part is the Kd-Tree structure corresponding to the 2D points in the left.

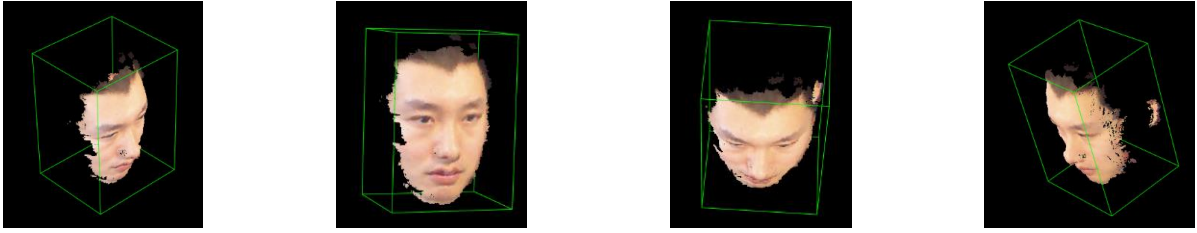

Fig. 5. Dense 3D facial point cloud obtained at the end of the final stage of the proposed method.

With obtained neighborhood of the points in the initial 3D facial point cloud, the average distance between seed points can be calculated easily. We use 6 times of average distance as the $R_0$ to perform RBF interpolation. The RGB colors of the interpolated points are the average color of its neighboring seed points. As shown in Figure 5, a denser point cloud of 3D face can be obtained with full details of the surface. As it can be observed, most holes are filled up, and the surface is much smoother. More experiments with other subjects are presented in the next section.

## 6. EXPERIMENTS AND RESULTS

The proposed method is implemented using C++, which is compiled by Microsoft Visual Studio. Microsoft's Kinect is used as demonstration to acquire the raw data. The initial depth image for the whole scene is under the resolution of VGA. There are totally six subjects in our experiments. Kinect only takes one single shot of each subject's face, and the proposed method successfully reconstructs the dense 3D point cloud of his/her face.

As shown in Figure 7 (a), it can be easily observed that data holes exist on the surface of the nose area in the initial 3D facial point cloud. With the final stage of our method, the majority of the holes in that area is filled up leading to a denser point cloud. The surface is continuous without the loss of details, as shown in Figure 7 (b). The six participants for experiments are shown in Figure 6 with their facial region recognized in green rectangles. The dense 3D facial point clouds reconstructed for them are illustrated in Figure 8. As shown in the results, our method can extract human 3D face regions exactly from the complex scenes. The details of the participant's face can be depicted in the reconstructed 3D facial model. The surface of the face is continuous, and almost all the data holes in the face are filled up. The performance of the proposed method is stable across six participants.

The density of the reconstructed point cloud is high enough to convey detailed information of a human face. The performance of the dense 3D point reconstruction is shown in Table 1, which contains the numbers of the recovered 3D points in two stages of the proposed method for comparison. According to Table 1, with the proposed method, the second stage can generate the points almost 10 times more than the points in the 3D point cloud obtained in the first stage. This makes the surface of the 3D face much smoother.

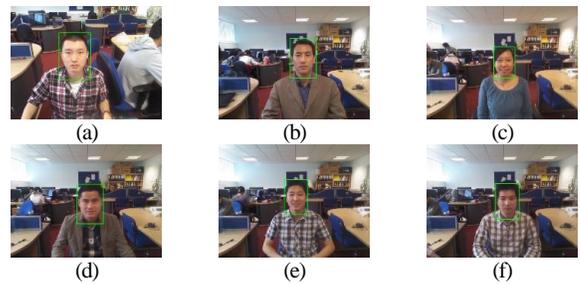

Fig. 6. Six participants' RGB image with recognized facial regions in green rectangles

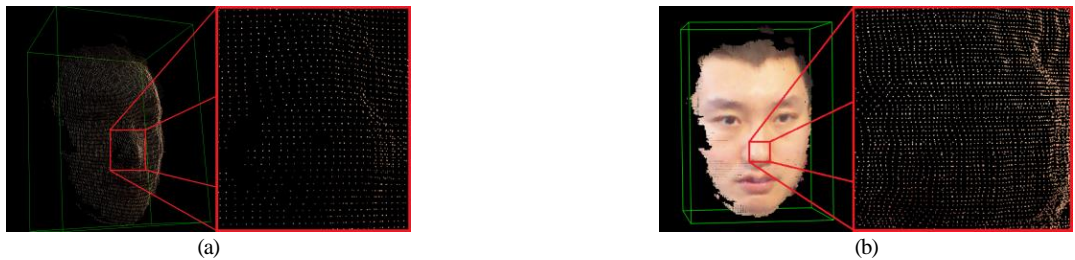

Fig. 7. Demonstration of the interpolation performance in the nose area. (**a**) The initial 3D facial point cloud achieved by the first stage; (**b**) The final dense 3D facial model obtained by the second stage. The nose area both in (**a**) and (**b**) is magnified for comparison.

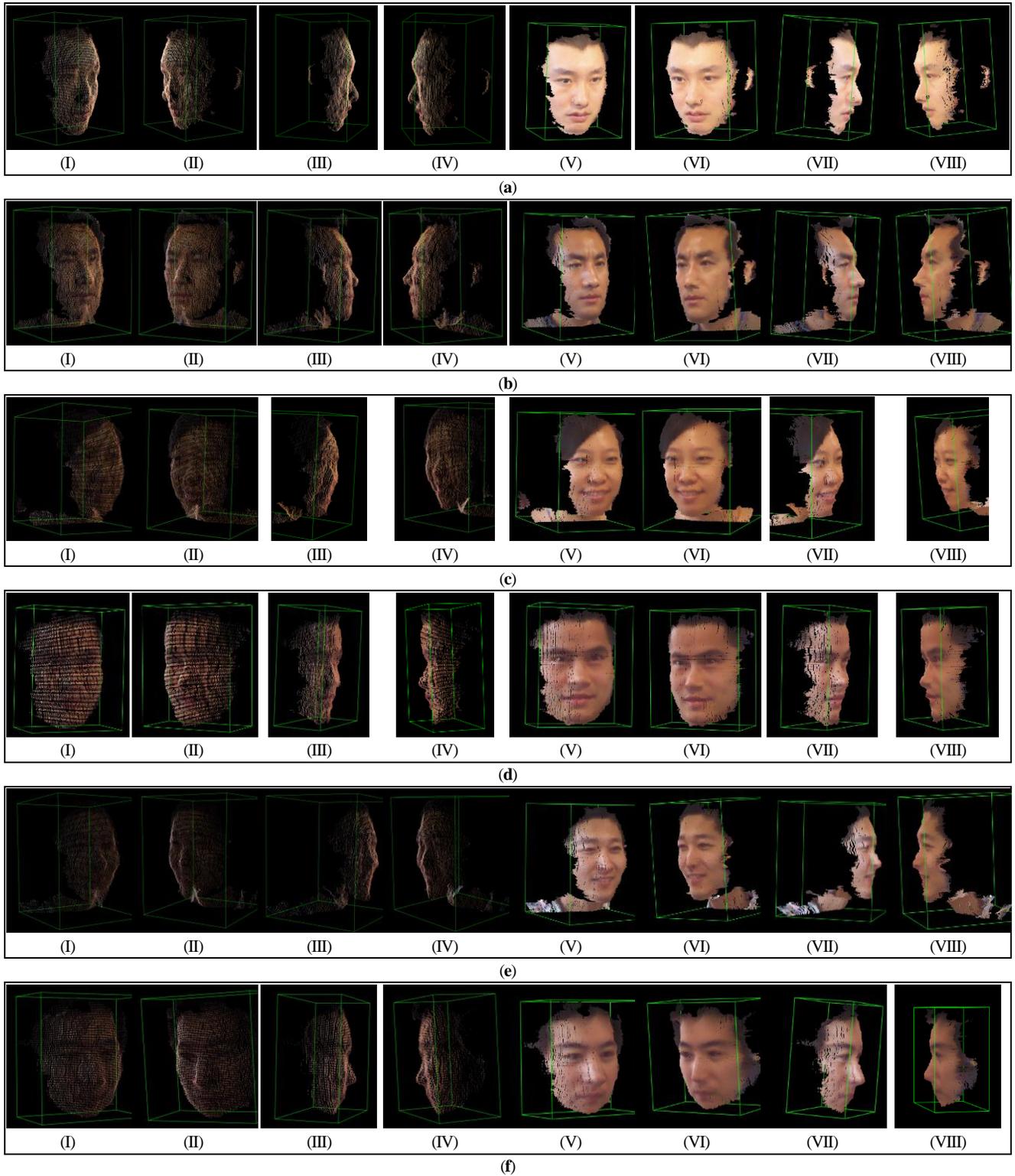

Fig. 8. Dense 3D facial point cloud producing results. (**a**) to (**f**) are six group of experiments with one subject in each group. In each group, (I) to (IV) are the initial 3D facial point cloud obtained after the first stage of proposed method, while (V) to (VIII) are the dense 3D facial point cloud obtained after final stage of the proposed method.

## Table 1. Interpolation Performance

| Participants [1] | Numbers of 3D Points Two Stages of Facial Point Cloud (FPC) | |
|---|---|---|
| | Initial FPC | Dense FPC |
| (a) | 12859 | 114658 |
| (b) | 13267 | 132476 |
| (c) | 9825 | 97137 |
| (d) | 9582 | 80374 |
| (e) | 9687 | 101326 |
| (f) | 8352 | 79482 |

[1] Participants are corresponding to the ones in Figure 6.

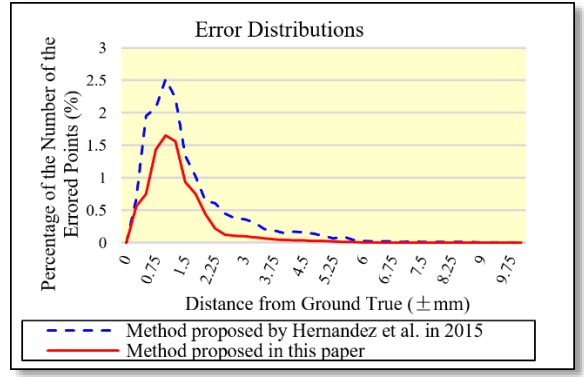

Fig. 9. Comparisons of error distributions

## Table 2. Comparisons of Means and Standard Deviations of the Errors.

| Methods | Data Errors (±mm) | |
|---|---|---|
| | Means | Standard Deviations |
| Method by Hernandez *et al.* in 2015 | 1.61 | 1.62 |
| Method by Hwang *et al.* in 2012 | 3.58 | 0.59 |
| Method by Lin *et al.* in 2002 | 3.12 | 1.14 |
| Method proposed in this paper | 1.31 | 1.07 |

We also compare our method with other similar methods proposed in the literatures for 3D facial modelling. As shown in Figure 9, the comparison is demonstrated for the percentages of the errored points in the point clouds generated using our method and the method proposed by Hernandez *et al.* in 2015 (Hernandez et al. 2015). As can be observed, almost no point in the point cloud has the error larger than ±5.5 mm in our method. The proposed method has a more stable performance with less errors in the reconstructed results. Moreover, over-smoothness in the results is successfully suppressed in our method, which is not the case for Hernandez's approach. Other comparisons are demonstrated in Table 2, which shows the competitive results of the proposed method.

## 7. CONCLUSION

In this paper, a novel method that generates dense 3D facial point cloud is presented. The first contribution of the proposed method is that we simplify the process of dense 3D facial reconstruction, which is easily achieved with only one shot of an off-the-shelf RGB-D sensor aimed at a human face without any further manual intervention. It is most suitable for the VR tasks that require to be carried out by the system itself. Moreover, the reconstructed point cloud has a smooth surface with detailed information of human face along with facial texture, which makes Virtual 3D facial analysis more accurate and more effective. The total number of the points in the cloud is around a hundred thousand. The second contribution is that the proposed method can be achieved at a very low cost without degradation of the performance. The only sensor used in our method is an inexpensive off-the-shelf RGB-D sensor, for example, the Microsoft's Kinect as demonstrated in this paper. Furthermore, the low resolution of the inexpensive sensor does not affect the performance of the proposed method. Applications of the proposed method can be expected in many face-related tasks in VR applications.

## Funding Information.

EU seventh framework programme under grant agreement No. 611391, Development of Robot-Enhanced Therapy for Children with Autism Spectrum Disorders (DREAM); Research Project of State Key Laboratory of Mechanical System and Vibration China MSV201508; National Natural Science Foundation of China (NSFC) (No. 41576011); International Science & Technology Cooperation Program of China (ISTCP) (N0. 2014DFA10410)